\documentclass[runningheads]{llncs}
\usepackage{graphicx}

\usepackage{tikz}
\usepackage{comment}
\usepackage{amsmath,amssymb} 
\usepackage{color}

\usepackage{algpseudocode}
\usepackage{booktabs}
\usepackage{tabularx}
\usepackage{array}
\usepackage{bm}
\usepackage{url}
\usepackage{bbm}
\usepackage{xcolor}
\usepackage{stackengine}
\usepackage{wrapfig}
\usepackage{multirow}
\usepackage{multicol}
\usepackage{tablefootnote}
\usepackage{comment}
\usepackage{makecell}
\usepackage{marvosym}
\usepackage[pagebackref,breaklinks,colorlinks]{hyperref}
\usepackage{cleveref}
\usepackage[width=122mm,left=12mm,paperwidth=146mm,height=193mm,top=12mm,paperheight=217mm]{geometry}

\newcommand\blfootnote[1]{%
\begingroup 
\renewcommand\thefootnote{}\footnote{#1}%
\addtocounter{footnote}{-1}%
\endgroup 
}
\begin{document}

\pagestyle{headings}
\mainmatter

\title{
GIMO: Gaze-Informed Human Motion Prediction in Context\\
} 

\author{Yang Zheng\inst{1,2} \,\,\,
Yanchao Yang\inst{1,}\textsuperscript{*} \,\,\,
Kaichun Mo\inst{1} \,\,\,
Jiaman Li\inst{1} \,\,\,
\\Tao Yu\inst{2} \,\,\,
Yebin Liu\inst{2} \,\,\,
C. Karen Liu\inst{1} \,\,\,
Leonidas J. Guibas\inst{1} \,\,\,
}
\authorrunning{Y. Zheng et al.}
\institute{
\,\inst{1} Stanford University \,\,\,\,\,\, \inst{2} Tsinghua University}

\maketitle

\begin{abstract}
    Predicting human motion is critical for assistive robots and AR/VR applications, where the interaction with humans needs to be safe and comfortable. 
    Meanwhile, an accurate prediction depends on understanding both the scene context and human intentions.
    Even though many works study scene-aware human motion prediction, the latter is largely underexplored due to the lack of ego-centric views that disclose human intent and the limited diversity in motion and scenes.
    To reduce the gap, we propose a large-scale human motion dataset that delivers high-quality body pose sequences, scene scans, as well as ego-centric views with the eye gaze that serves as a surrogate for inferring human intent.
    By employing inertial sensors for motion capture, our data collection is not tied to specific scenes, which further boosts the motion dynamics observed from our subjects. 
    We perform an extensive study of the benefits of leveraging the eye gaze for ego-centric human motion prediction with various state-of-the-art architectures.
    Moreover, to realize the full potential of the gaze, we propose a novel network architecture that enables bidirectional communication between the gaze and motion branches.
    Our network achieves the top performance in human motion prediction on the proposed dataset, thanks to the intent information from eye gaze and the denoised gaze feature modulated by the motion.
    Code and data can be found at \href{https://github.com/y-zheng18/GIMO}{https://github.com/y-zheng18/GIMO}.
    \end{abstract}
\blfootnote{\textsuperscript{*} Corresponding author: yanchaoy@cs.stanford.edu}
\begin{figure}[ht]
    \centering
    \includegraphics[width=\linewidth]{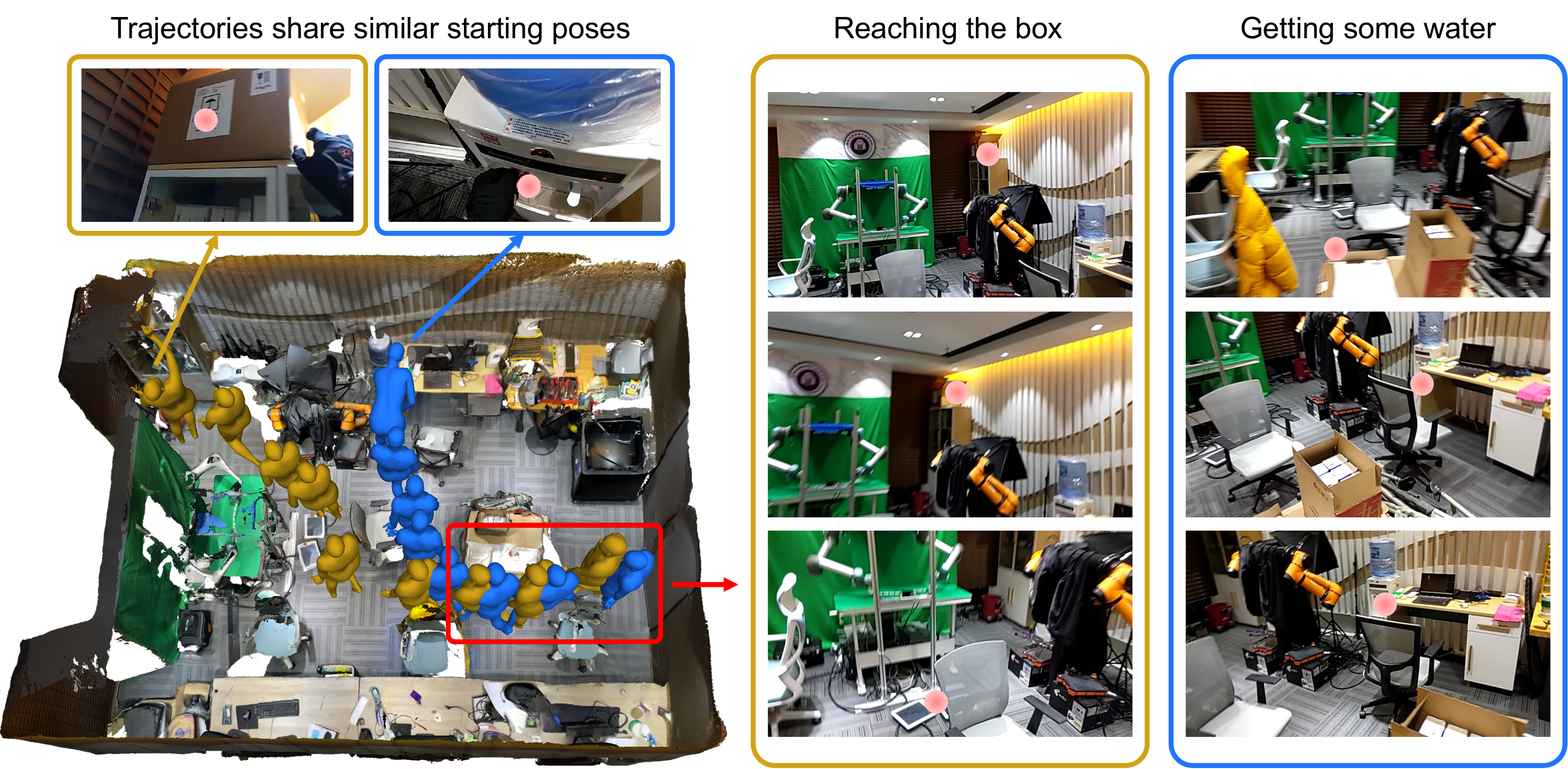}
    \vspace{-10pt}
    \caption{Human motion driven by different intents look similar at the beginning. However, the scanning patterns of the eye gaze (red dots) during the starting phase are pretty distinctive, which suggests that we can leverage eye gaze to reduce uncertainties when predicting future body movements.}
    \label{fig:gaze_different_tasks}
    \vspace{-15pt}
\end{figure}

\section{Introduction}

A large portion of the human brain cortex is devoted to processing visual signals collected by the optic nerve, and over half of the nerve fibers carry information from the fovea that is responsible for sharp central vision. 
When modulated through foveal fixation, or equivalently, {\it eye gaze,} important sensory input of fine details perceived with the fovea can inform future actions of the human agent \cite{tatler2011eye,gottlieb2013information}.
As shown in Fig.~\ref{fig:gaze_different_tasks},
a human agent intending to perform two tasks entails distinctive gaze patterns, even though the first few moves are not very distinguishable.
Hence, it is beneficial to employ eye gaze when making human motion predictions in the 3D scene, which is of great importance for human-machine interactions \cite{admoni2017social,duarte2018action}.
For example, a human agent wearing an AR/VR headset may approach a chair to sit on it or just grab a cup on the table behind it.
If the latter is true, we may want the headset to send out a warning for collision avoidance based on the forecast future.
To resolve ambiguities for reliable human motion prediction,
there is an increasing interest in leveraging eye gaze 
as it highly correlates to the underlying intent that motivates the consequent actions.

The key to understanding the role of gaze and how it can effectively inform human motion prediction lies in two folds.
First, it is critical to have a dataset with high-quality 3D body pose annotations and corresponding eye gaze information.
Besides data quality, the 3D scene and motion dynamics should be diverse to enable meaningful learning and evaluation of the gain when eye gaze is incorporated.
Second, it is also crucial to have a network architecture that can efficiently utilize {\it sparse} eye gaze during predictions given the multi-modal setting (e.g., gaze, human motion and scene geometry) and the fact that {\it not} every single gaze is of the {\it same} significance regarding the agent's intent (e.g., one may get distracted by a salient object in the scene that has nothing to do with the task at hand).

However, most existing human motion datasets {\it do not} support evaluating the effect of eye gaze due to the lack of ego-centric data annotated with both gaze and 3D body pose within the same scene. 
Recently, there are a few datasets proposed on ego-centric social interaction and object manipulation where gaze and the viewer's 3D poses are available.
Nevertheless, they are not suitable for ego-centric human motion prediction since the diversity of scenes and the variation in motion dynamics are very limited.
To validate the benefits of eye gaze in {\it human motion prediction},
we propose a large-scale ego-centric dataset, which contains the scene context, eye gaze, and accurate 3D body poses of the human actors.
By employing an advanced motion capture system based on Inertial Measurement Units (IMUs), we can collect 3D pose data with high fidelity and avoid the limits of conventional multi-camera systems.
For example, the actor can walk through any environment without performing a cumbersome setup of motion capture devices.
Moreover, accurate poses can be recorded without any 2D-3D lifting, which could induce errors due to occlusions and noise in the detection.
These advantages enable the actors to perform various long-horizon activities in a diverse set of daily living environments.

In order to check the effectiveness of eye gaze in improving human motion prediction, we perform an extensive study with multiple state-of-the-art architectures.
However, we note that gaze and motion could both be inherently ambiguous in forecasting future movements. 
For example, the gaze may be allocated to a TV monitor while walking towards the dining table.
In this case, the actor may simply follow the momentum, thus rendering the eye gaze uninformative about the body motion.
To utilize the full potential of eye gaze in human motion prediction,
we further propose a novel architecture that manifests cross-modal attention such that
{\it not only} future motion can benefit from the eye gaze, {\it but also} the significance of gaze in predicting the future can be reinforced by the observed motion. 
In our experiments, better human motion predictions are observed across various architectures.
Furthermore, the proposed architecture achieves the top performance measured under different criteria, verifying the effectiveness of our bidirectional fusion scheme. 

In summary, we make the following contributions.
First, we provide a large-scale human motion dataset that enables investigating the benefits of eye gaze under diverse scenes and motion dynamics.
Second, we propose a novel architecture with a bidirectional multi-modal fusion that better suits gaze-informed human motion prediction through mutually disambiguating motion and gaze.
Finally, we validate the usefulness of eye gaze for human motion prediction with multiple architectures and verify the effectiveness of the proposed architecture by showing top performance on the proposed dataset.

\section{Related Work}

\vspace{-5pt}{\bf Datasets for human motions.}
Human motion modeling is a long-standing problem and is extensively explored with high-quality motion capture datasets, ranging from small-scale CMU Graphics Lab Motion Capture Database~\cite{cmu_mocap} to large-scale ones like AMASS~\cite{mahmood2019amass}. Human3.6M~\cite{ionescu2013human3} captures high-quality motions using a multi-view camera system and serves as a standard benchmark for motion prediction and 3D pose estimation.   
While these datasets provide adequate data to learn motion dynamics, the constraints from the 3D environment are usually not included. 
Later, more datasets containing the 3D scene are proposed, and scene-aware motion prediction can be studied using GTA-1M dataset~\cite{cao2020long}. PROX~\cite{hassan2019resolving} includes both 3D scene and human interaction motions which can be used to explore scene-aware motion generation~\cite{wang2021synthesizing} task and the problem of placing human to the scene~\cite{zhang2020place,zhang2020generating}.
As the data is always collected with a human agent, ego-centric videos are provided in EgoPose~\cite{yuan2019ego,yuan20183d}, Kinpoly~\cite{luo2021dynamics} and HPS~\cite{guzov2021human} to study how the motion estimation and prediction can benefit from these ego-centric observations.
Moreover, social interaction is considered in You2Me~\cite{ng2020you2me} and EgoBody \cite{zhang2021egobody}. 
However, existing datasets do not contain diverse 3D scenes and human motions with intentions, we collect a large-scale dataset for gaze-guided human motion prediction, and it consists of high-quality human motions, 3D scene, ego-centric video and corresponding eye gaze information.

\vspace{1mm}\noindent {\bf Human motion prediction.}
RNNs have proven successful in modeling human motion dynamics~\cite{fragkiadaki2015recurrent,martinez2017human,li2017auto,aksan2019structured}.
~\cite{mao2020history} proposes an attention-based model to guide the future prediction with motion history. To effectively exploit both spatial and temporal dependencies in human pose sequences, ST-Transformer~\cite{aksan2021spatio} designs a spatial temporal transformer architecture to 
model the human motions. Pose Transformers~\cite{martinez2021pose} investigates a non-autoregressive formulation using transformer model and shows superior performance in terms of both efficiency and accuracy. As human motions are tightly correlated with the scene context, scene-aware motion prediction is also actively studied~\cite{cao2020long,hassan2021stochastic}. A three-stage pipeline is established to predict long-term human motions conditioned on the scene context~\cite{cao2020long}. SAMP~\cite{hassan2021stochastic} further includes object geometry to estimate interaction positions and orientations, and generates motions following a calculated collision-free trajectory. Besides the scene constraints, other modalities such as gaze and music also provide clues for future motion prediction. Transformer~\cite{vaswani2017attention} is applied to generate dance movements conditioned on music~\cite{li2020learning,li2021ai,valle2021transflower}. MoGaze~\cite{kratzer2020mogaze} verifies the effectiveness of eye gaze information for motion prediction with an RNN model in a full-body manipulation scenario. Our work aims to predict long-term future motions with both 3D scene and gaze constraints. We differ from existing motion prediction works, as their focus is the dense motion predictions, while we are predicting long-term sparse motions to understand human intentions.

\vspace{1mm}\noindent {\bf Human motion estimation.} 
3D pose estimation is extensively studied in third-person view images or videos~\cite{tian2022hmrsurvey,zhang2021pymaf,kocabas2021pare,kolotouros2019learning,kocabas2020vibe,luo20203d,hossain2018exploiting}. VIBE~\cite{kocabas2020vibe} propose a sequential model to estimate human poses and shapes from videos, along with a motion discriminator to constrain the predictions in a plausible motion manifold.
TCMR~\cite{hossain2018exploiting} explicitly enforces the neural nets to leverage past and future frames to eliminate jitters in predictions. Motion priors are founded effective in improving the temporal smoothness and tackling the occlusion issues~\cite{rempe2021humor,li2021task,zhang2021learning}. 
More attentions are received in ego-centric pose estimation recently. Pose estimation from images captured using a fish eye camera is explored in ~\cite{tome2019xr,xu2019mo,rhodin2016egocap,tome2020selfpose,wang2021estimating}. ~\cite{jiang2017seeing} deploy a chest-mounted camera and predict motions based on an implicit motion graph. Following the chest-mounted camera setting, You2Me~\cite{ng2020you2me} introduces the motions of the visible second person as an additional signal to constrain the motion estimation of the camera wearer. ~\cite{yuan2019ego,yuan20183d,luo2021dynamics} explores motion estimation and prediction with head-mounted front-facing camera. 
In this work, we are addressing the ego-centric motion prediction task where past motions are given. Our proposed dataset can benefit the ego-centric motion estimation problem.

\begin{figure}[!t]
    \vspace{-5pt}
    \centering
    \includegraphics[width=\linewidth]{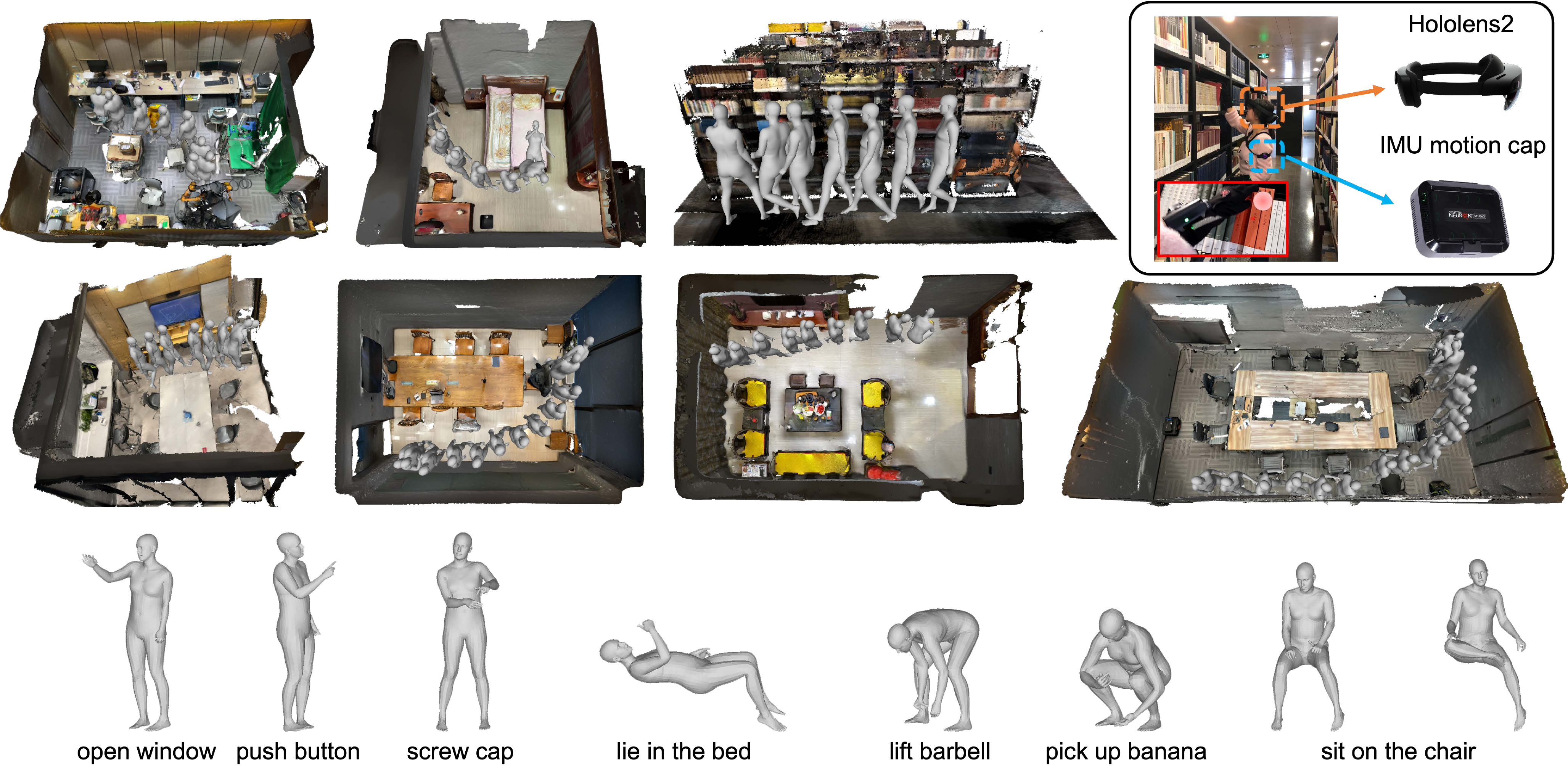}
    \vspace{-15pt}
    \caption{We collect human motion data in various indoor environments (1st, 2nd rows), allowing the human subject to perform a diverse range of daily activities exhibiting rich dynamics (bottom). Top-right: motion and gaze capture devices.}
    \label{fig:teaser}
    \vspace{-15pt}
\end{figure}

\section{GIMO Dataset: Gaze and Motion with Scene Context}

\noindent Human motion is affected by the scene, which provides physical constraints and the agent's psychological demand that drives body movements.
To have a concrete assessment of the benefits induced by eye gaze, we need both ego-centric views, and 3D body poses of the agent.
Particularly, they should be temporally synchronized and spatially aligned within the 3D scenes.
Current datasets for human motion prediction are either collected in a virtual environment risking being unrealistic or 
captured by an array of cameras with limited scene diversity and motion dynamics. Moreover, eye gaze is usually not available.

Therefore, we propose a real-world large-scale dataset 
that provides high-quality human motions, ego-centric views with eye gaze, as well as 3D environments. Next, we describe our data collection pipeline.

\subsection{Hardware Setup}

We employ a commercialized IMU-based motion capture system to record high-quality 3D body poses of the human agent, whose eye gaze in 3D is detected using an AR device mounted on the head.
The 3D scenes are scanned by a smartphone equipped with lidar sensors (please see Fig.~\ref{fig:teaser}, top-right).

\vspace{0.5mm}
\noindent\textbf{Motion capture.} 
To capture daily activities in various indoor environments, 
we resort to motion capture from IMU signals following HPS~\cite{guzov2021human}.
While HPS only provides SMPL~\cite{loper2015smpl} models with body movements, 
we take advantage of an advanced commercial product \emph{Noitom PERCEPTION NEURON STUDIO,}\footnote{https://noitom.com/perception-neuron-series}
which can record at 96 fps 3D body and hand joint movement of the subject. 
To obtain the full-body pose and hand gesture of the subject,
we apply SMPL-X~\cite{pavlakos2019expressive} model to fit the recorded IMU signals from multiple joints.
Compared to human motion datasets like PROX~\cite{hassan2019resolving}, where the 3D body pose is estimated from monocular RGB videos, the pose obtained using the above procedure is free from estimation errors caused by noisy detection and occlusions.
Fitting parametric human body models for poses from multi-view RGB(D) streams or with marker-based systems is also commonly used to collect human motion data \cite{joo2018total,zhang2021egobody,ionescu2013human3},
however, 
our pipeline requires much less effort in presetting the environment; thus, we can collect human motion data in any indoor scene.
These characteristics endow us with the capability to ensure the diversity of the scene and motion dynamics in our dataset.

\newcolumntype{P}[1]{>{\centering\arraybackslash}p{#1}}
\begin{table}[!t]
    \vspace{-5pt}
    \caption{Statistics of existing and our datasets. 
$^*$ means the 3D scene is virtual, e.g., from game engine~\cite{cao2020long} or CAD models~\cite{hassan2021stochastic}. 
\emph{Ego} denotes egocentric images are available, and \emph{Intent} indicates whether the motions have clear semantic intentions, e.g., fetching a book.}
    \label{tab:stats}
    \vspace{-7pt}
    \begin{center}
        \resizebox{\linewidth}{!}{
        \begin{tabular}{cP{1cm}P{1cm}cP{1cm}P{1cm}cccP{1.5cm}c}
            \toprule
            Dataset & Frame & Sub. & 3D scene & Ego & Gaze & 3rd-person & \makecell[c]{Human pose\\from} & \makecell[c]{Parametric\\model} & Intent
             & \makecell[c]{Task} \\  
            \midrule 
            EGTEA Gaze+~\cite{li2021eye} & 2419k & 32 & & \checkmark & \checkmark & & & & \checkmark & Action recognition\\	
            TIA~\cite{wei2018and} & 330k & - & & & \checkmark & \checkmark & & & \checkmark & Attention prediction\\
            Human3.6M~\cite{ionescu2013human3} & 3600k & 11 & & & & \checkmark & Marker-based & & & Pose estimation \\ 
            TNT15~\cite{von2016human} & 13k & 4 & & & & \checkmark & RGB+IMU & & & Pose estimation \\
            3DPW~\cite{von2018recovering} & 51k & 7 & & & & \checkmark & RGB+IMU & SMPL & & Pose estimation \\
            Panoptic~\cite{joo2015panoptic} & 297k & 180+ & & & & \checkmark & Multi-RGB & & & Pose estimation \\
            TotalCapture~\cite{joo2018total} & 1,900k & 5  & & & & \checkmark & Multi-RGB & Frank & & Pose estimation \\
            HPS~\cite{guzov2021human} & 300k & 7 & \checkmark & \checkmark & & & IMU & SMPL & & Pose estimation\\
            EgoBody~\cite{zhang2021egobody} & 153k & 20 & \checkmark & \checkmark & \checkmark & \checkmark & Multi-RGB-D & SMPL-X & & Pose estimation\\
            EgoMoCap~\cite{luo2021dynamics} & 148k & 3 &  & \checkmark &  & & Marker-based & & & Pose estimation\\
            PROX~\cite{hassan2019resolving},~\cite{zhang2020generating} & 100k & 20 & \checkmark & & & \checkmark & RGB & SMPL-X & & Human generation\\
            GTA-IM~\cite{cao2020long} & 1000k & & \checkmark$^*$ & & & \checkmark & Game engine & & & Motion prediction\\
            SAMP~\cite{hassan2021stochastic} & & 1 & \checkmark$^*$ & & & & Marker-based & SMPL-X & \checkmark & Motion prediction\\
            \midrule
            GIMO (ours) & 129k & 11 & \checkmark & \checkmark & \checkmark &  & IMU & SMPL-X & \checkmark & Motion prediction \\
            \bottomrule
    \end{tabular}}
    \end{center}
    \vspace{-25pt}
\end{table}

\vspace{0.5mm}
\noindent\textbf{Gaze capture.} 
Following~\cite{zhang2021egobody}, we use Hololens2\footnote{https://www.microsoft.com/en-us/hololens} and its Research Mode API~\cite{ungureanu2020hololens} to capture the 3D eye gaze. 
It also records ego-centric video at 30 fps in $760\times 428$ resolution, long-throw depth streams at 1-5 fps in $512\times 512$, and 6D poses of the head-mounted camera. 
The 3D scene is reconstructed through TSDF fusion given the recorded depth, which is used for the subsequent global alignment. The eye gaze is recorded as a 3D point in the coordinate system of the headset.

\vspace{0.5mm}
\noindent\textbf{3D scene acquisition.} 
To obtain high-quality 3D geometry of the scene (the reconstructed TSDF result from Hololens2 is usually noisy), we use an iPhone13 Pro Max equipped with LiDAR sensors to scan the environment through 3D Scanner APP\footnote{https://apps.apple.com/us/app/3d-scanner-app/id1419913995}. 
The output mesh contains about 500k vertices and photorealistic texture, providing sufficient details to infer the affordance of the scene. 
The data collection process involving human agents and the alignment of different coordinate frames to the scanned meshes are described in the following.

\subsection{Data Collection with Human in Action}

One distinct feature of our dataset is that it captures long-term motions with clear intentions. 
Different from prior datasets for motion estimation purposes where the subjects are performing random actions such as jumping and waving hands, we aim at collecting motion trajectories with semantic meaning, e.g., walk to open the door.
Thus, we focus on collecting data from various daily activities in indoor scenes.
The full statistics of our dataset are listed in Tab.~\ref{tab:stats}.

\begin{table}[!t]
    \vspace{-5pt}
    \caption{Activities performed by our subjects.}
    \vspace{-5pt}
    \label{tab:activities}
    \begin{center}
        \resizebox{0.8\linewidth}{!}{
        \begin{tabular}{cc}
            \toprule
            Category & Activities \\  
            \midrule 
            Resting & Sitting or laying on objects\\
            Interacting with objects & \makecell[c]{Touching, holding, stepping on, reaching to objects}\\
            Changing the state of objects & \makecell[c]{Opening, pushing, transferring, throwing, \\
                                                        picking up, lifting, connecting, screwing, \\
                                                        grabbing, swapping objects
                                                          }\\
            \bottomrule
    \end{tabular}}
    \end{center}
    \vspace{-25pt}
\end{table}

To this end, we recruit 11 university students (4 female and 7 male) and ask them to perform the activities defined in Tab.~\ref{tab:activities}.
The subjects are instructed to start from a distant location to the goal object and then move to the destination to act.
Therefore, long-term motion with clear intention can be obtained. Especially, the collection progress includes the following steps: (i) the subject wears the head-mounted Hololens2, the IMU-based motion capture clothes, and gloves, where calibration is performed to set up the motion capture system; 
(ii) the subject chooses the action from the activities in Tab.~\ref{tab:activities} according to the affordance of the scene; 
(iii) the 3D scene is scanned; 
(iv) the subject starts to carry out the planned activities in the scene while data are collected; 
(v) the scene is reset for the following subjects to perform their activities.
Note, if the subject changes the scene geometry, we reset the objects to their original states to avoid rescanning the whole environment.

As a result, our dataset contains 129k ego-centric images, 11 subjects, and 217 motion trajectories in 19 scenes, manifesting enough capacity and diversity for gaze-informed human motion prediction. 
As illustrated in Fig.~\ref{fig:teaser}, the motions are smooth and convey clear semantic intentions.

\subsection{Data Preparation}

\noindent\textbf{Synchronization.}
Given compatibility issues, it is difficult to synchronize the motion capture system with Hololens2 without modifying their commercialized software.
Instead, we use a hand gesture that can be observed in the ego-centric view as a starting signal. 
Once the pose and ego-centric image of the hand gesture are aligned, the rest frames can be synchronized according to the timestamps. 

\vspace{0.5mm}
\noindent\textbf{Parametric model fitting.} 
To obtain the 3D body pose and shape of the subject,
we fit SMPL-X~\cite{pavlakos2019expressive} model to the 3D joints (23 body joints, 15 left-hand joints, and 15 right-hand joints), which are computed from the recorded IMU signals by the provided commercial software. 
In addition, the 6D head pose is used to determine the head position and orientation of the SMPL-X model.

\vspace{0.5mm}
\noindent\textbf{Alignment.} 
The Hololens2 coordinate system and the fitted SMPL-X models need to align with the high-quality 3D scene scans. 
The former is aligned through ICP between the TSDF fusion result of the depth recorded by Hololens2 and the 3D scene. 
The SMPL-X motion sequence is first transformed to the Hololens2 coordinate system via human annotations, i.e., the start and end shapes of the human body are scanned by Hololens2 and visible in the TSDF reconstruction, which serves as anchor shapes for aligning the fitted models.
The pose can then be aligned to the 3D scene using the global transformation obtained from the previous ICP alignment between the scene scans.
Our dataset is named GIMO, and we describe our method for gaze-informed motion prediction in the following.

\section{Gaze-Informed Human Motion Prediction}
\label{sec:method}
\begin{figure}[!t]
    \centering
    \includegraphics[width=\linewidth]{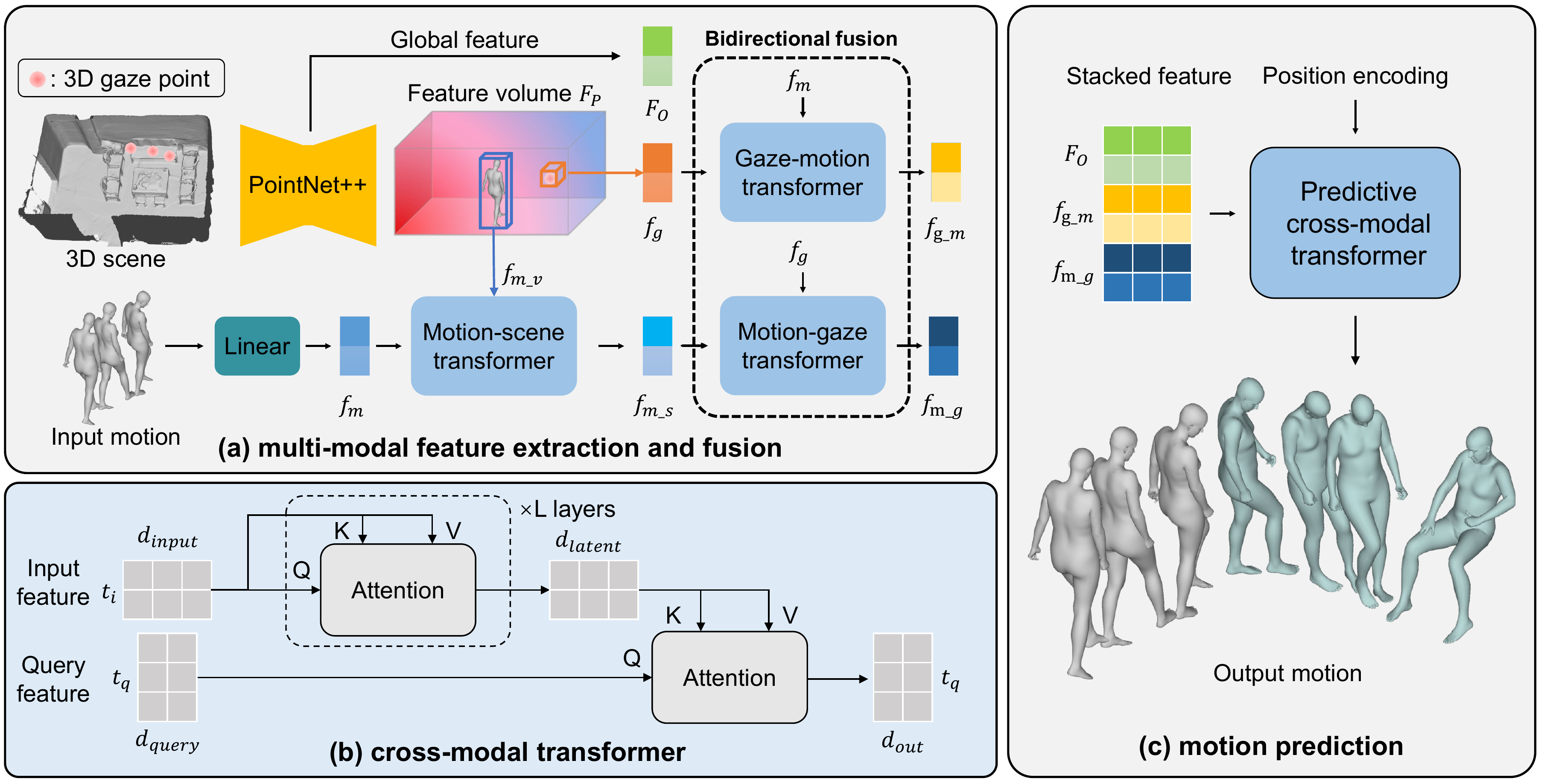}
    \vspace{-10pt}
    \caption{Our gaze-informed human motion prediction architecture. Multi-modal features, i.e., gaze feature, human motion feature, and global scene feature, are extracted and then fused through the proposed {\it bidirectional} fusion scheme (a). The fused features are then stacked into a holistic representation and used for future motion prediction (c). The cross-modal transformer component~\cite{jaegle2021perceiver} is illustrated in (b). Please refer to Sec.~\ref{sec:method} for more details.}
    \label{fig:network}
    \vspace{-15pt}
 \end{figure}

Gaze conveys relevant information about the subject's intent, 
which can be used to enhance long-horizon human motion prediction. 
On the other hand, past motions~\cite{cao2020long,aksan2021spatio}, ego-centric views~\cite{yuan2019ego,hassan2021stochastic}, or 3D context~\cite{hassan2021stochastic,wang2021synthesizing} could provide helpful constraints on human motion, yet, the prediction is still challenging and suffers from uncertainties in the future. 
Here, we aim at gaze-informed long-term human motion prediction. 
Specifically, given the past motion, 3D scene, and 3D eye gaze as inputs, we study how they can be integrated to resolve the ambiguities in future motion and generate intention-aware motion predictions.

To fully utilize the geometry information provided by the 3D scene and intention clues from past motions and gaze, we propose a novel framework with a bidirectional fusion scheme that facilitates the communication between different modalities. 
As shown in Fig.~\ref{fig:network}, we use PointNet++~\cite{qi2017pointnet++} as the encoding backbone to extract per-point features of the 3D scene, followed by several cross-modal transformers to transcend information from multi-modality embeddings.

\subsection{Problem Definition}

We represent a motion sample as a parametric sequence $X_{i:j}=\{x_i, x_{i+1}, \cdots, x_{j}\}$ where $x_{k}=(t_k, r_k, h_k, \beta_{k}, p_{k})$ is a pose frame at time $k$. 
Here $t\in R^3$ is the global translation, $r\in SO(3)$ denotes the global orientation, $h_k\in R^{32}$ refers to the body pose embedding, $\beta\in R^{10}$ is the shape parameter, 
and $p\in R^{24}$ is the hand pose, where SMPL-X body mesh $M=\mathcal{M}(t_k, r_k, h_k, \beta_{k}, p_{k})$ can be obtained using VPoser~\cite{pavlakos2019expressive}. 
The 3D scene is represented as a point cloud $S\in R^{n\times 3}$, and the 3D gaze point $g\in R^3$ is defined as the intersection points between the gaze direction and the scene.
Thus, given the inputs of a motion sequence $X_{1:t}$ along with the corresponding 3D gaze $G_{1:t}=\{g_1, g_2, \cdots, g_t\}$ and the 3D scene $S$,
we aim to predict the future motion $X_{t:t+T}=\Phi(X_{1:t}, G_{1:t}, S|\theta)$ where $\theta$ represents the network parameters.

\subsection{Multi-modal Feature Extraction}
Instead of extracting the multi-modal embeddings independently~\cite{li2021ai}, we propose a novel scheme to integrate the motion, gaze, and scene features. 
The gist is to let the motion and gaze features communicate to each other, so their uncertainties regarding the future can be mutually decreased, resulting in more effective utilization of the gaze information.

\vspace{0.5mm}
\noindent\textbf{Scene feature extraction.}
To learn the constraints from the 3D scene and guide the network to pay attention to local geometric structures, 
we apply PointNet++ to extract both global and local scene features. 
Specially, we obtain the per-point feature map and a global descriptor of the scene as follows:
\begin{equation} 
\label{eq:pointnet++}
F_P, F_o = \Phi_{scene}(S|\theta_{s})
\end{equation}
where $S\in R^{n\times 3}$ is the input point cloud, $F_P\in R^{n\times d_p}$ is the per-point $d_p$ dimensional feature map, and $F_o\in R^{d_o}$ is the global descriptor of the scene. 
Given the per-point feature $F_P$, the feature of an arbitrary point $e$ can be computed through the inversed distance weighted interpolation~\cite{qi2017pointnet++}:
\begin{equation} 
    \label{eq:feats volume}
F_{P|e} = \frac{\Sigma_{i=1}^{n_e}w_iF_{P|p_i}}{\Sigma_{i=1}^{n_e}w_i}, w_i = \frac{1}{||p_{i} - e||_2}
\end{equation}
where $\{p_{1}, p_{2}, \cdots, p_{n_e}\}$ are the nearest neighbors of $e$ in the scene point cloud.

\vspace{0.5mm}
\noindent\textbf{Gaze feature extraction.}
\label{sec:gaze feats}
We query the gaze point feature $f_g$ from the per-point scene feature map $F_P$ according to Eq.~\ref{eq:feats volume}, i.e., $f_g=F_{P|g}$.
Thus, the interpolated gaze feature contains relevant scene information that provides cues to infer the subject's intention.

\vspace{0.5mm}
\noindent\textbf{Motion feature extraction.} A linear layer is used to extract the motion embedding $f_m$ from the input motion parameter $x$. 
To endow the embedding awareness of the 3D scene, we further query the scene features of the SMPL-X vertices using Eq.~\ref{eq:feats volume}.
These SMPL-X per-vertex features are then fed to PointNet~\cite{qi2017pointnet} to get the ambient scene context feature $f_{m\_v}$ of the current motion pose:
\begin{equation} 
    \label{eq:f_m_v}
f_{m\_v} = PointNet(\{F_{P|v}, v\in \mathcal{M}(x)\}\})
\end{equation}
where $\mathcal{M}(x)$ is the SMPL-X vertex set with motion parameter $x$.

\subsection{Attention-Aware Multi-modal Feature Fusion}
Given the multi-modal nature of the gaze, scene, and motion, an efficient feature fusion module is necessary to leverage the information from different modalities.
Instead of directly concatenating the features~\cite{li2021ai}, we propose a more effective scheme by deploying a cross-modal transformer~\cite{jaegle2021perceiver} to fuse the gaze, motion, and scene features (Fig.~\ref{fig:network}). 
We explain our design in the following.

\vspace{0.5mm}
\noindent\textbf{Cross-modal transformer.}
The cross-modal transformer~\cite{jaegle2021perceiver} is used to capture the correlations between input embedding sequences and to establish communications between the multi-modal information. 
It is largely based on attention mechanism~\cite{vaswani2017attention}. 
An attention function~\cite{jaegle2021perceiver} maps a query and key-value pairs to an output as:
\begin{equation}   
\label{eq:attention}
Attention(Q,K,V)=softmax(\frac{QK^T}{\sqrt{d_K}})V, Q=qW_q, K=kW_k, V=vW_v
\end{equation}
where $q\in R^{l_q\times d_q}$, $k\in R^{l_{kv}\times d_k}$, $v\in R^{l_{kv}\times d_v}$ are input query, key and value vectors, and $W_q\in R^{d_q\times d_K}$, 
$W_k\in R^{d_k\times d_K}$, $W_q\in R^{d_v\times d_V}$ embed the inputs. Here $d$ denotes the dimension of the input vector and $l$ is the sequence length. 

As shown in Fig.~\ref{fig:network} (b), the cross-modal transformer is built on a stack of attention layers, which maps a $t_i$-length input into a $t_q$-length output by querying a $t_q$-length feature:
\begin{equation}   
    \label{eq:cross_att}
    \phi_{out} = cross\_trans(\phi_{query}, \phi_{input})
    \end{equation}
It is proved to be efficient in processing multi-modal signals, e.g., text, \& audio. 

\vspace{0.5mm}\noindent\textbf{Motion feature fusion.}
The motion feature should be aware of the 3D scene context and the subject's intention inferred from the gaze information,
so that it can guide the prediction network to generate more reasonable motion trajectories (e.g., free from penetration and collision) and accurate estimations of the ending position or pose of the subject. 
For this purpose, we first use the scene context feature $f_{m\_v}$ acquired from the ambient 3D environment (Eq.~\ref{eq:f_m_v}) as the query to update the motion feature $f_m$ through a motion-scene transformer:
\begin{equation}
    \label{eq:f_m_s}
    f_{m\_s} = cross\_trans(f_{m\_v}, f_m)
\end{equation}
Thus, the output motion embedding $f_{m\_s}$ is expected to be aware of the 3D scene. 
We then feed $f_{m\_s}$ to the next motion-gaze transformer where the gaze feature $f_g$ is the query input: 
\begin{equation}
    \label{eq:f_m_g}
    f_{m\_g} = cross\_trans(f_g, f_{m\_s})
\end{equation}
The final motion embedding $f_{m\_g}$ is expected to integrate both the 3D scene information and the intention clues from the gaze features.

\vspace{0.5mm}\noindent\textbf{Gaze feature fusion.}
While gaze can help generate intention-aware motion features, the motion could also provide informative guidance to mitigate the randomness of gaze since not every gaze point reveals meaningful user intent. 
Therefore, we treat the gaze embedding in a bidirectional manner, i.e., the motion embedding $f_m$ is also used as the query to update the gaze features such that the network can learn which gaze features contribute more to the future motion:
\begin{equation}
    \label{eq:f_g_m}
    f_{g\_m} = cross\_trans(f_m, f_g)
\end{equation}
The bidirectionally fused multi-modal features are then composed into holistic temporal representations of the input to perform human motion prediction.
As illustrated in Fig.~\ref{fig:network} (c),
the updated gaze feature $f_{g\_m}$, motion feature $f_{m\_g}$ and the global scene feature $F_O$ are used to predict the future motion by: 
\begin{equation}
    \label{eq:out}
X_{t:t+T} = cross\_trans(h_{position}, cat(f_{g\_m}, f_{m\_g}, F_O)_{1:t})
\end{equation}
where $cat$ denotes the concatenation operation, and $h_{position}$ is the latent vector that contains temporal positional encodings for the output~\cite{jaegle2021perceiver}. We verify the effectiveness of our design in utilizing gaze information through experiments.

\section{Experiments}
\begin{figure}[t!]
    \centering
    \includegraphics[width=\linewidth]{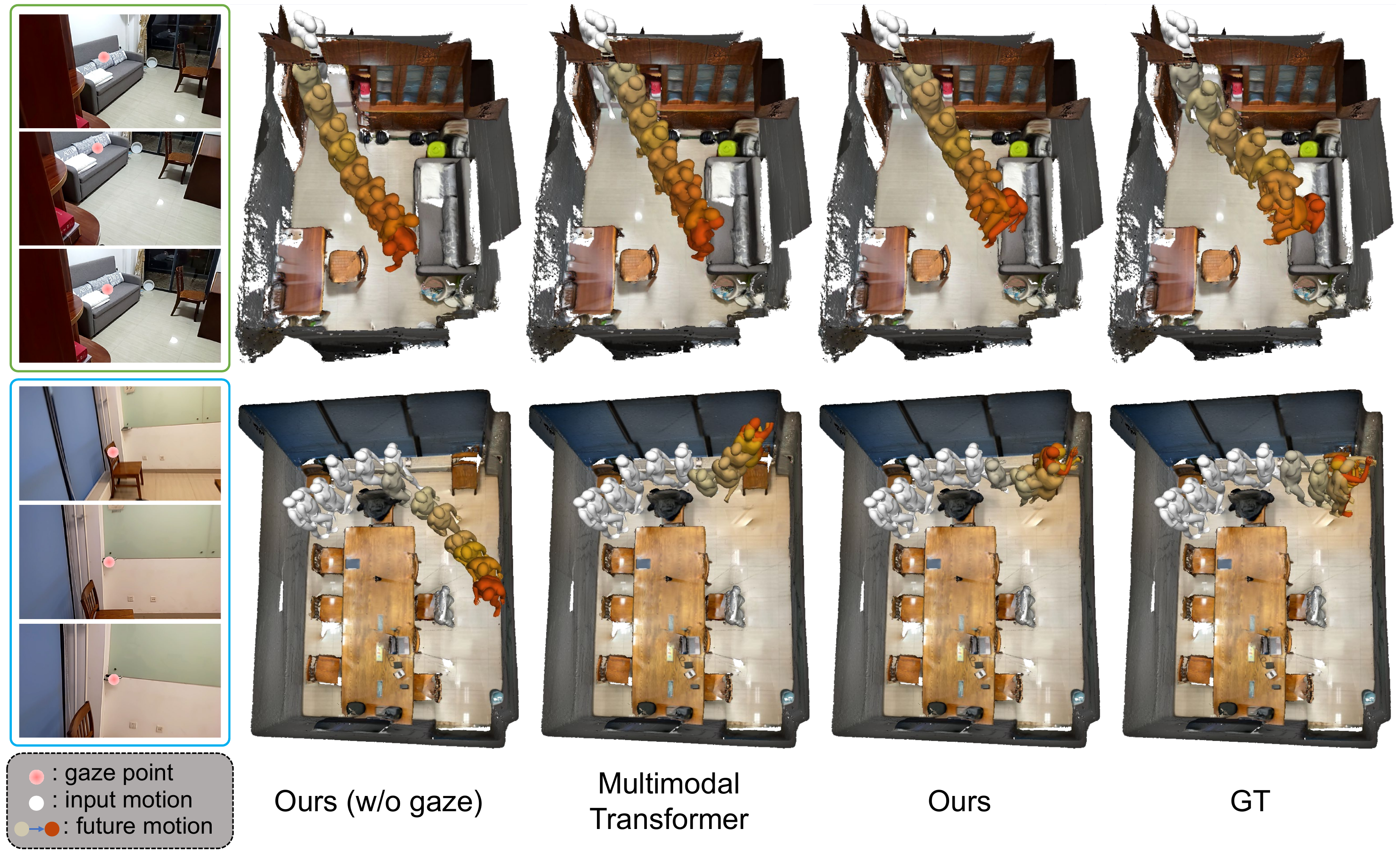}
    \vspace{-10pt}
    \caption{Qualitative results. Top row: results on a known scene from the training set. Bottom row: results in a new environment. We compare our method with MultimodalNet~\cite{li2021ai} and ours without gaze. Please zoom in for details.}
    \vspace{-15pt}
    \label{fig:results}
\end{figure}
In this part, we explain our experimental setup and results. Our goal is to examine the following questions: 
\begin{enumerate}
    \item Does gaze help disambiguate human motion prediction?
    \item How do state-of-the-art methods perform on our dataset? 
    \item What is the contribution of each part of our design to the final performance? Overall, is the proposed architecture effective?
\end{enumerate}

\subsection{Experimental Setup}
In our experiments, we predict the future motion in 5 seconds from 3 seconds input, where the first 3 seconds of a trajectory is just about to start an activity (i.e., beginning to move for fetching a book) in our dataset, and in the next 5 seconds the trajectory proceeds to finish the activity. 
We set the motion frame rate to 2 fps, i.e., 6 pose input and 10 pose output. 
Note that once the waypoints are predicted, a full motion sequence with high fps can be easily generated~\cite{wang2021synthesizing}. 
Since we aim to explore the effect of gaze in disambiguating motion prediction, high-frequency motion is not necessary in our experiments. 

\vspace{0.5mm}\noindent\textbf{Baselines.} 
We implement several state-of-the-art motion prediction and generation baselines including spatio-temporal transformer~\cite{aksan2021spatio} and an RNN network~\cite{kratzer2020prediction} for full motion prediction from the past motion input, 
and MultimodalNet~\cite{li2021ai} based on transformer for motion synthesis from multi-modal data (i.e., gaze, motion, and the 3D scene feature in our experiments). 
We build our pipeline by incorporating 6 cross-modal transformer layers~\cite{jaegle2021perceiver} to extract 256-dimensional gaze and motion features. L1 loss between the predicted motion and the ground truth is used to train the network. More details about the network architecture and training are available in the supplementary material. 

\begin{table}[b!]
    \vspace{-25pt}
    \caption{Destination accuracy. We report the global translation and orientation error and mean per-joint position error (\emph{MPJPE}).}
    \vspace{-5pt}
    \label{tab:des error}
    \begin{center}
        
        \resizebox{0.8\linewidth}{!}{
        \begin{tabular}{P{4cm}P{1cm}P{1cm}P{1.5cm}P{1cm}P{1cm}P{1.5cm}}
            \toprule
            \multirow{2}*{Method} & \multicolumn{3}{c}{\makecell[c]{Known scenes}} & \multicolumn{3}{c}{\makecell[c]{New scenes}}\\ 
            & \tiny{Trans} & \tiny{Ori} & \tiny{MPJPE} & \tiny{Trans} & \tiny{Ori} & \tiny{MPJPE}  \\
            \midrule
            ST-Transformer~\cite{aksan2021spatio} & 0.587 & 0.864 & 279.9 & 0.516 & 0.682 & 236.8\\
            RNN~\cite{kratzer2020prediction} & 0.538 & 0.822 & 272.5& 0.547 & 0.894& 230.4\\
            MultimodalNet~\cite{li2021ai} & 0.442 & 0.699 & 260.0 & 0.389 & 0.658 & 236.0 \\
            \midrule
            RNN+gaze~\cite{kratzer2020mogaze} & 0.389 & 0.882 & 264.2 & 0.345 & 0.611 & 230.0 \\
            \makecell[c]{MultimodalNet+gaze~\cite{li2021ai}} & 0.316 & 0.743 & 266.6 & 0.300 & 0.583 & \textbf{204.9} \\
            \midrule
            Ours \footnotesize{(w/o gaze)} & 0.393 & 0.656 & 262.1 & 0.389 & 0.709 & 228.7\\
            Ours \footnotesize{(pointnet)} & 0.310 & 0.659 & 240.6 & 0.394 & 0.563 & 234.5\\
            Ours \footnotesize{(vanilla)} & 0.353 & 0.739 & 249.0 & 0.365 & 0.602 & 220.4\\
            Ours & \textbf{0.245} & \textbf{0.579} & \textbf{237.8} & \textbf{0.280} & \textbf{0.556} & 209.0\\
            \bottomrule
    \end{tabular}}
    \vspace{-10pt}
    \end{center}
\end{table}

\begin{table}[t!]
    \vspace{-10pt}
    \caption{Path errors of the predicted motions.}
    \vspace{-5pt}
    \label{tab:path error}
    \begin{center}
        
        \resizebox{0.8\linewidth}{!}{
        \begin{tabular}{P{4cm}P{1cm}P{1cm}P{1.5cm}P{1cm}P{1cm}P{1.5cm}}
            \toprule
            \multirow{2}*{Method} & \multicolumn{3}{c}{\makecell[c]{Known scenes}} & \multicolumn{3}{c}{\makecell[c]{New scenes}}\\ 
            & \tiny{Trans} & \tiny{Ori} & \tiny{MPJPE} & \tiny{Trans} & \tiny{Ori} & \tiny{MPJPE}  \\
            \midrule
            ST-Transformer~\cite{aksan2021spatio} & 0.329 & 0.503 & 201.4 & 0.339 & 0.537 & 201.7\\
            RNN~\cite{kratzer2020prediction} & 0.308 & 0.476 & 195.2& 0.324 & 0.495 & 180.3\\
            MultimodalNet~\cite{li2021ai} & 0.273 & 0.383 & 190.0 & 0.294 & 0.425 & 177.0 \\
            \midrule
            RNN+gaze~\cite{kratzer2020mogaze} & 0.235 & 0.457 & 190.1 & 0.278 & \textbf{0.288} & 182.6 \\
            \makecell[c]{MultimodalNet+gaze~\cite{li2021ai}} & 0.246 & 0.424 & 193.1 & \textbf{0.250} & 0.374 & 183.7 \\
            \midrule
            Ours \footnotesize{(w/o gaze)} & 0.305 & 0.412 & 180.1 & 0.315 & 0.403 & 182.5\\
            Ours \footnotesize{(pointnet)} & 0.218 & 0.360 & 180.5 & 0.267 & 0.403 & 184.5\\
            Ours \footnotesize{(vanilla)} & 0.238 & 0.399 & 182.7 & 0.286 & 0.348 & 180.3\\
            Ours & \textbf{0.213} & \textbf{0.340} & \textbf{177.1} & 0.261 & 0.322 & \textbf{160.3}\\
            \bottomrule
    \end{tabular}}
    \end{center}
    \vspace{-15pt}
\end{table}

\subsection{Evaluation}
To evaluate, we divide the 217 trajectories of our dataset into 180 trajectories for training and 37 for testing. 
The 37 motions consist of 27 trajectories (different from the training ones) performed in known scenes from the training set and 10 in 2 new environments scanned only for evaluation purposes.

\begin{figure}[t!]
    \centering
    \includegraphics[width=0.8\linewidth]{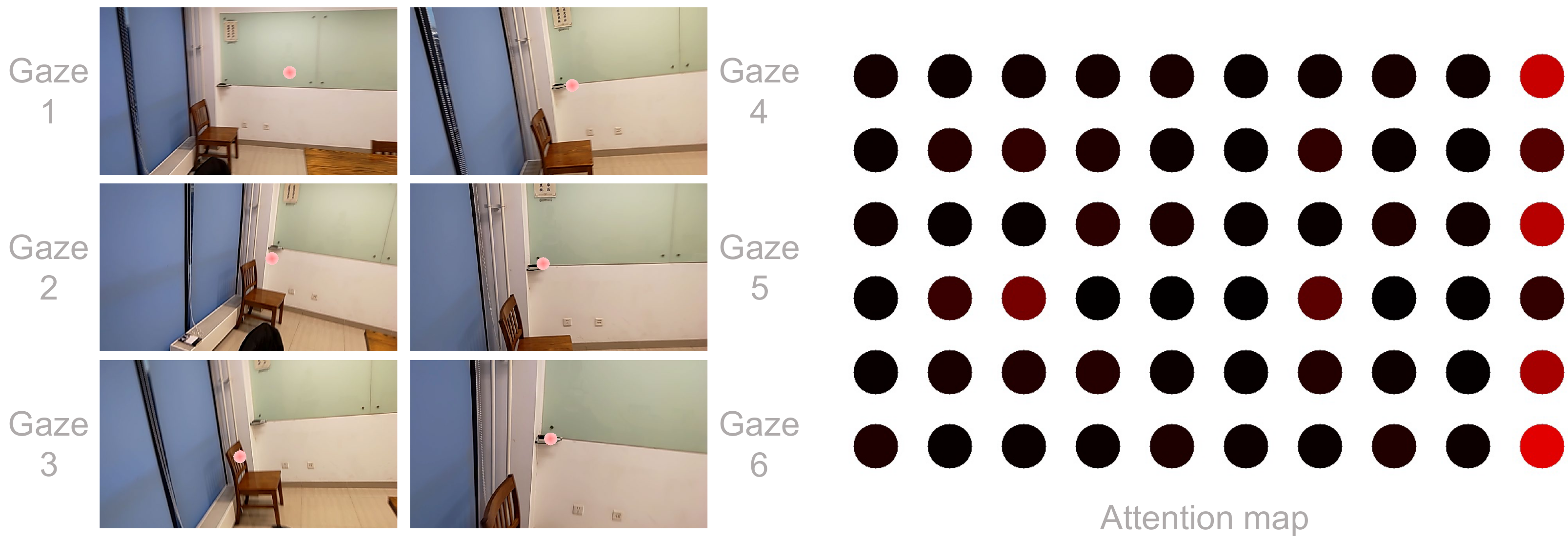}
    \vspace{-5pt}
    \caption{The attention map of the 6 input gaze for the 10 output motion. The gaze influences the ending output most (brighter means larger weight), indicating that the gaze features reveal the subject's final goals.}
    \label{fig:atts}
    \vspace{-15pt}
\end{figure}
\vspace{0.5mm}\noindent\textbf{Evaluation metrics.} 
We employ the destination error and the path error as our evaluation metrics. 
The destination error refers to the global translation, rotation error and the mean per-joint position error (MPJPE)~\cite{ionescu2013human3} of the last pose in the predicted motion. 
The destination pose contains essential information about the subject's goal, which is our experiments' primary focus. 
The path error is computed as the mean error of the predicted poses in 5s~\cite{cao2020long}.
We treat the global translation and rotation error as the $l1$ distance between the predicted SMPL-X translation and orientation parameter with the ground truth~\cite{wang2021synthesizing}.
 
\vspace{0.5mm}\noindent\textbf{Quantitative evaluation.} 
As shown in Tab.~\ref{tab:des error} and Tab.~\ref{tab:path error}, while the state-of-the-art methods based on spatio-temporal transformer~\cite{aksan2021spatio} suffer from ambiguities since the prediction is simply from the past motion, a simple RNN method with motion and gaze input~\cite{kratzer2020mogaze} can significantly reduce the ambiguity, indicating the effectiveness of gaze in guiding the prediction of motion.
Our method achieves promising results in predicting reasonable future motion with small destination and translation errors. 
Compared to MultimodalNet~\cite{li2021ai} built on the vanilla transformers~\cite{valle2021transflower}, our method outperforms in recognizing the subject's intent from the gaze and thus predicts more accurate destination poses.

\vspace{0.5mm}\noindent\textbf{Qualitative evaluation.} 
Fig.~\ref{fig:results} shows that in a ''going to sit'' activity performed in one scene from the training set (top row), 
our method manages to generate accurate destination poses, i.e., sitting on the sofa. 
In the new environment, the subject first grabs a blackboard eraser and then starts wiping. 
While all the methods generate walking actions, ours without gaze input fails to predict the correct motion. 
When given gaze, results from MultimodalNet~\cite{li2021ai} and our method both reach out the hand and try to grab something. 
Our prediction successfully arrives at the destination point where the eraser lies; however, the results of MultimodalNet~\cite{li2021ai} reach out to the wrong place.
More visualizations and failure cases are included in the supplementary material.

\subsection{Ablation Study}
In this part, we aim to answer question 3 by finding the factors that contribute to the superior performance of our method.

\vspace{0.5mm}\noindent\textbf{Variant 1: gaze.} 
We evaluate the baseline's performance with and without gaze input to explore how gaze could influence the motion prediction results. 
As clearly demonstrated in Tab.~\ref{tab:des error} and Tab.~\ref{tab:path error}, 
the RNN network~\cite{kratzer2020mogaze} and the MultimodalNet~\cite{li2021ai} both gain significant accuracy improvement given gaze inputs. 
Fig.~\ref{fig:results} shows that without gaze, our method is confused about the future destination. 
To find more intuitions about the role of gaze in motion prediction, we visualize the attention weights of gaze feature query over the motion feature as depicted in Fig.~\ref{fig:atts}.
Interestingly, we find the gaze feature does influence the ending poses in the predicted motion, implying that the gaze can serve as a strong indicator of the destination of a motion, which reveals the user's intent.

\vspace{0.5mm}\noindent\textbf{Variant 2: pointnet++ for scene feature query.} 
We propose to use Pointnet++~\cite{qi2017pointnet++} to extract the per-point feature of the scene such that the gaze feature and scene-aware motion feature can be obtained (\cref{sec:gaze feats}). 
We replace it with Pointnet to extract the global scene feature and use a linear layer to get the gaze feature. 
Results in Tab.~\ref{tab:des error} and Tab.~\ref{tab:path error} demonstrate that the variant can act well on scenes from the training set, but lose its competitiveness when generalized to new environments with different 3D structures.

\vspace{0.5mm}\noindent\textbf{Variant 3: cross-modal transformer.} 
The cross-modal transformer architecture proves to be effective in bridging multi-modal information~\cite{jaegle2021perceiver}. We replace it with the vanilla transformer~\cite{vaswani2017attention} as used in~\cite{li2021ai}. 
Results in Tab.~\ref{tab:des error} and Tab.~\ref{tab:path error}(\emph{Ours (vanilla)}) demonstrate the loss of accuracy compared to the full design. 
Note that the path error of the variant on the new scenes is even larger than the results without gaze input, indicating that the vanilla transformer might not be efficient enough to capture the correlations between multi-modal inputs. Thus a more sophisticated design such as a cross-modal transformer is needed.

\section{Conclusion, Discussion and Future Work}

We present the GIMO dataset, a real-world dataset with ego-centric images, 3D gazes, 3D scene context, and ground-truth human motions.
With the collected dataset, we define a new task, i.e., gaze-informed human motion prediction, which leverages eye gaze to infer the subject's potential intention to minimize the ambiguities in motion prediction.
We further contribute a novel framework, which achieves promising results in predicting long-term future motions. 
While our method only relies on 3D inputs, we aim as future work to incorporate visual information from ego-centric images to further boost the accuracy.

Instead of the proposed task, our dataset can benefit various applications, e.g., intention-aware motion synthesis and gaze-guided ego-centric pose estimation. 
We believe our work not only opens new directions for motion prediction but will have foreseeable impacts on ego-centric vision topics.

\paragraph{\bf Acknowledgments} 
The authors are supported by a grant from the Stanford HAI Institute, a Vannevar Bush Faculty Fellowship, a gift from the Amazon Research Awards program, the NSFC grant No.62125107, and No.62171255. Also, Toyota Research Institute provided funds to support this work.
{\bibliographystyle{splncs04}
\bibliography{egbib}
}

\setcounter{section}{0}
\renewcommand\thesection{\Alph{section}}
\newcommand{\suppsection}{\subsection}
\begin{flushleft}
\textbf{\Large Appendix}
\end{flushleft}
\makeatletter
\section{Experimental Setup}

\subsection{Implementation Details}
As demonstrated in the Fig. 3 of the main paper, our method is built on Pointnet++~\cite{qi2017pointnet++} and a cross-modal transformer~\cite{jaegle2021perceiver}. 
A 256D global feature $F_O$ and a 256D per-point feature map $F_P$ of the scene are extracted from the input point cloud. 
The feature of an arbitrary point $e$ is computed through the inversed distance weighted interpolation on the 3 nearest neighbors of $e$ from the scene point cloud (Eq. 2 of the main paper), where we query the 256D gaze feature $f_g$ and obtain the 256D scene context feature $f_{m\_v}$ of the current motion from SMPL-X per-vertex features. 
The 32D motion parameter $x$ is embedded into 256D motion feature $f_m$ through a linear layer. 
The motion embedding is then fed to a motion-scene transformer with $f_{m\_v}$ as query and further fed to another motion-gaze transformer with gaze feature $f_g$ as the query. 
The gaze feature is updated by a gaze-motion transformer queried by motion feature $f_m$. 
We then concatenate the global scene feature $F_O$, the updated motion feature $f_{m\_g}$ and gaze feature $f_{g\_m}$ to get the 768D multi-modal embedding, which is used to predict the 32D future motion parameter by a cross-modal transformer.
All the transformers adopt a 6 layer architecture as proposed in~\cite{jaegle2021perceiver} with 256D latent embedding.
Note that here the input and output motion parameter $x$ consists of a 3D global translation vector $t$, a 3D global orientation vector $r$ (represented as axis angle), and a 32D pose embedding $h$ obtained from VPoser~\cite{pavlakos2019expressive}. 
We omit predicting the hand poses $p$ and the shape parameter $\beta$ since the global body pose can be well represented by parameter $\{t, r, h\}$, and we aim at future work to include hand poses and the body shape for more detailed motion prediction.

For the baseline methods, we re-implement spatio-temporal transformer~\cite{aksan2021spatio}, a RNN based network~\cite{kratzer2020prediction}, and MultimodalNet~\cite{li2021ai} to adapt for our experimental settings. 
The 3D joint angle representation is used as motion input and output to train the spatio-temporal transformer and RNN as introduced in~\cite{kratzer2020prediction}, while MultimodalNet is based on the 32D motion parameter the same as ours. 
An 8 layer transformer~\cite{vaswani2017attention} with 512 embedding size and 8 heads attention is used in spatio-temporal transformer~\cite{aksan2021spatio}. 
A three layer RNN with 1024 hidden size is deployed to predict future motion with simple motion input or motion and gaze input~\cite{kratzer2020mogaze}. 
In MultimodalNet~\cite{li2021ai}, the motion input is firstly embedded into 256D feature space through linear layers and then fed to a transformer encoder to get the motion embeddings. The gaze embedding is also obtained with linear layers and a transformer encoder.
The global scene feature from PointNet~\cite{qi2017pointnet}, the gaze embedding and the motion embedding are stacked and fed to a transformer decoder to generate future motion. The transformer encoders and decoder are all based on a 6 layer architecture with 256 latent size. 
Therefore, all the baselines share similar network capacity with our method.

\begin{figure}[t!]
    \centering
    \includegraphics[width=\linewidth]{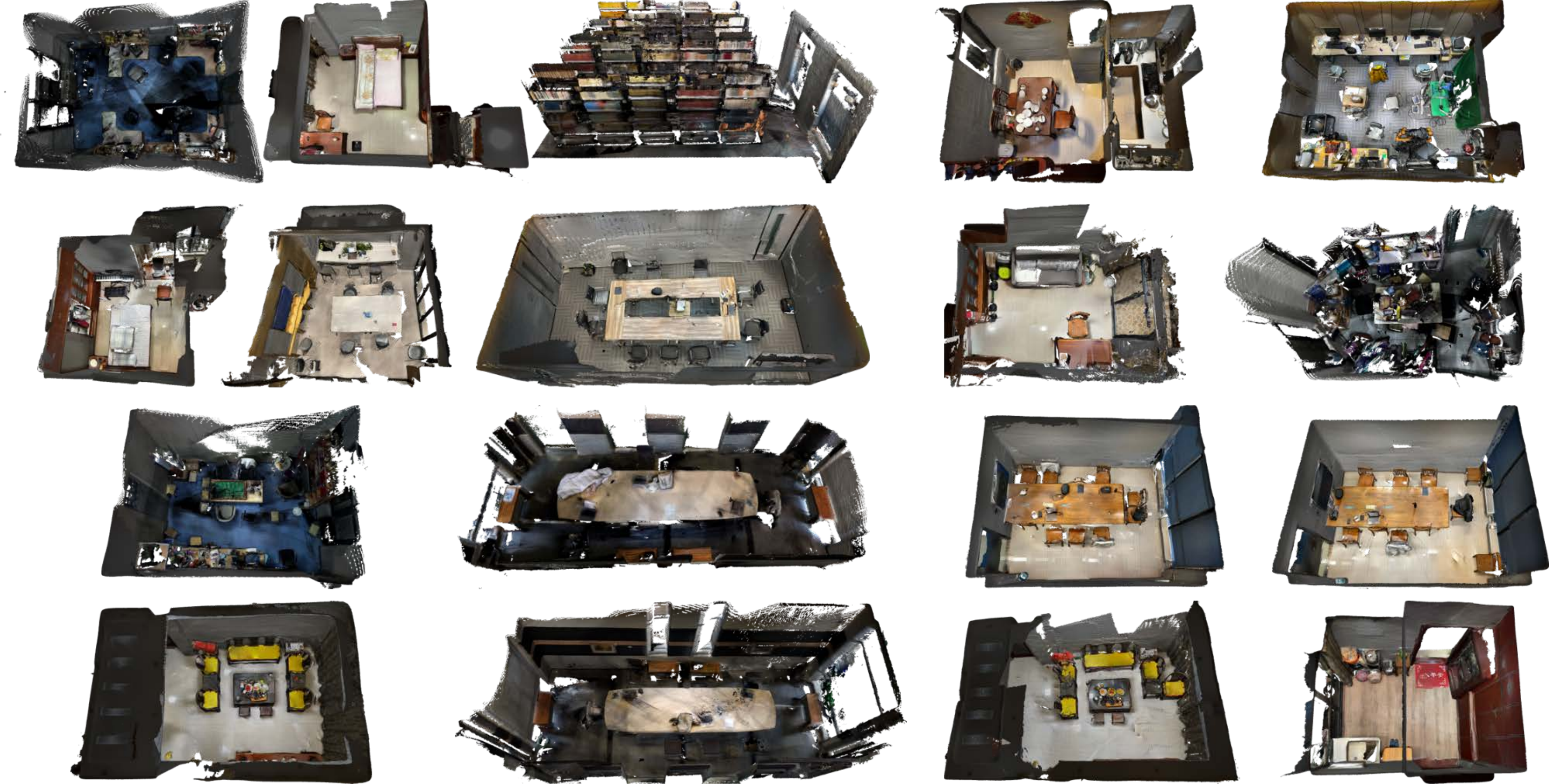}
    \caption{An overview of the scanned scenes in our dataset.}
    \label{fig:scenes}
 \end{figure}
\begin{figure}[t!]
    \centering
    \includegraphics[width=0.8\linewidth]{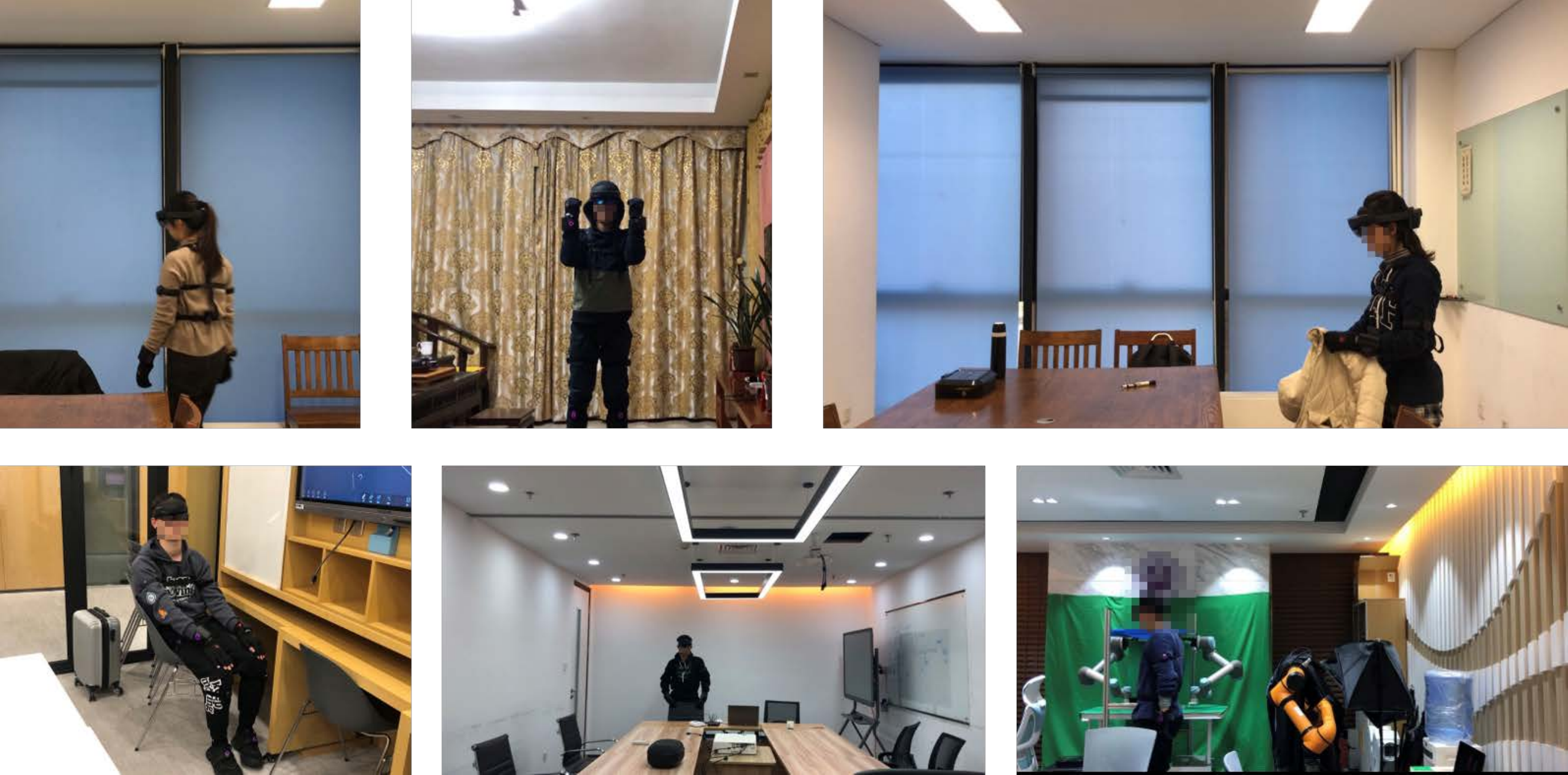}
    \caption{Subjects in the scenes.}
    \label{fig:sub}
 \end{figure}
\begin{figure}[t!]
    \centering
    \includegraphics[width=\linewidth]{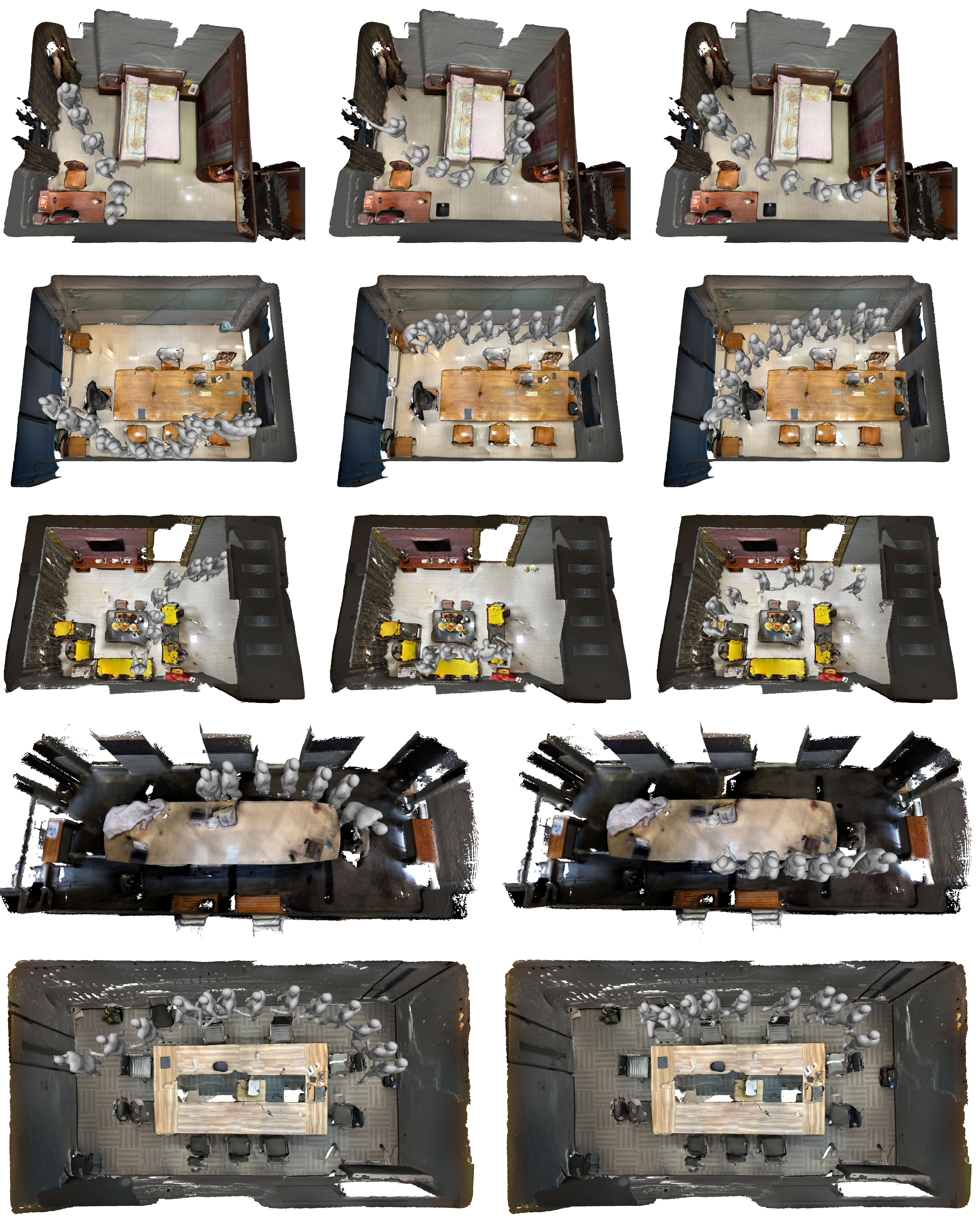}
    \caption{Motion trajectories from our dataset. Better visualized in the supplementary video.}
    \label{fig:sup motion}
 \end{figure}
\subsection{Training Loss}
We employ the L1 loss between the predicted motion parameter and ground truth to train our method. 
The full loss consists of translation loss, orientation loss and pose embedding loss.
The translation loss is formulated as:
\begin{equation}
    \label{eq:loss_t}
    \mathcal{L}_{trans} = \frac{1}{T}\mathop{\Sigma}_{k=1}^T||\widehat{t_k} - t_k||_1
\end{equation}
where $T$ is the length of output pose, and $\widehat{t_k}$ is the predicted global translation parameter of the k-th pose in the T-length future motion, and $t_k$ is the ground truth. 
We compute the orientation loss as:
\begin{equation}
    \label{eq:loss_o}
    \mathcal{L}_{ori} = \frac{1}{T}\mathop{\Sigma}_{k=1}^T||\widehat{r_k} - r_k||_1
\end{equation}
where $\widehat{r_k}$ is the predicted global orientation parameter. The pose embedding loss is designed as:
\begin{equation}
    \label{eq:loss_p}
    \mathcal{L}_p = \frac{1}{T}\mathop{\Sigma}_{k=1}^T||\widehat{h_k} - h_k||_1
\end{equation}
where $\widehat{h_k}$ is the predicted pose embedding. Finally, the full loss is formulated as:
\begin{equation}
    \label{eq:loss_f}
    \mathcal{L} = \lambda_t \mathcal{L}_{trans} + \lambda_o \mathcal{L}_{ori} + \lambda_p \mathcal{L}_{p}
\end{equation}
where we set $\lambda_t, \lambda_o, \lambda_p$ to 1 during training.
\begin{figure}[t!]
    \centering
    \includegraphics[width=\linewidth]{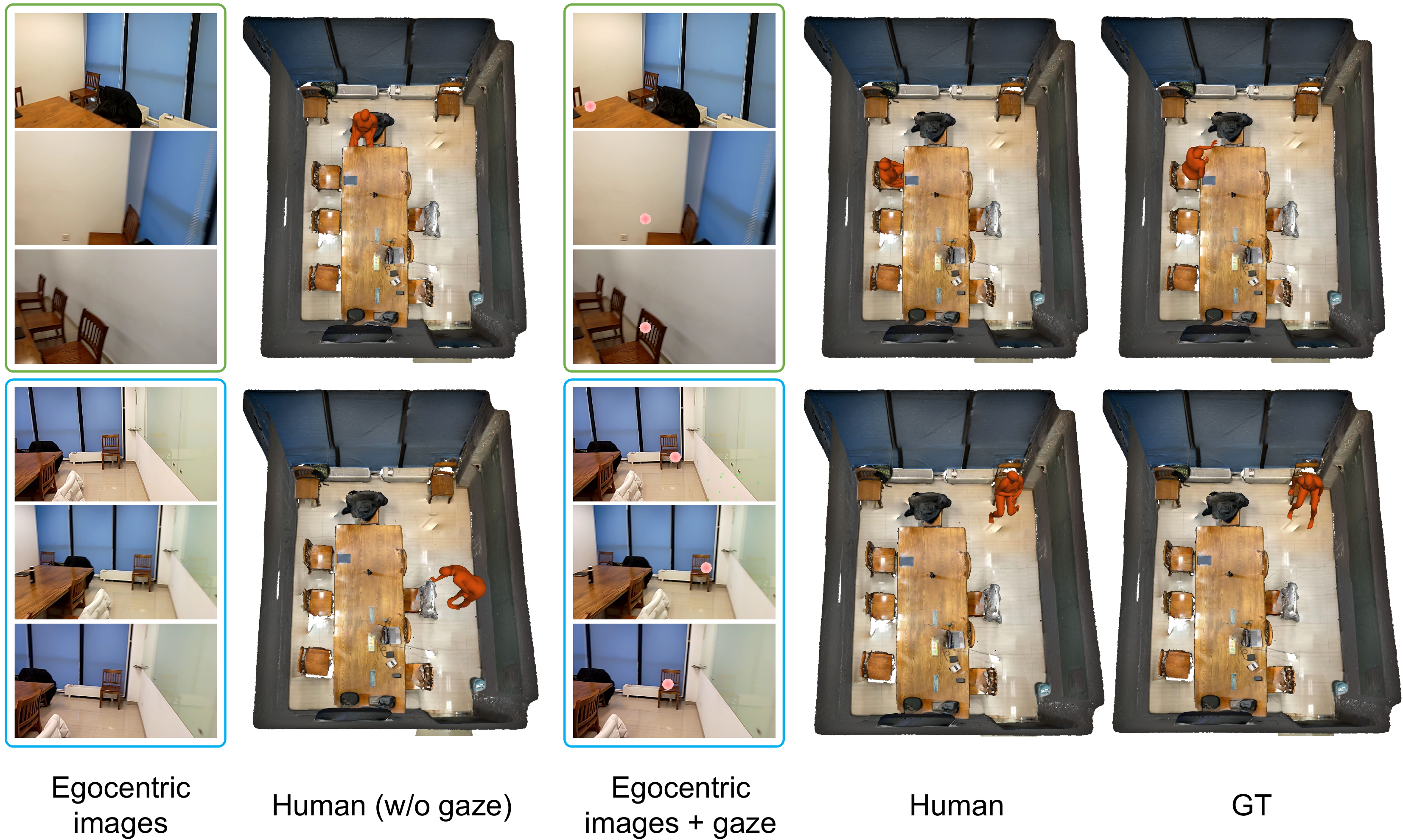}
    \caption{Human evaluation. Two human subjects are required to watch a egocentric video (without gaze or with gaze) and infer the final pose of the trajectory. The subjects choose a pose from a pose database which comes from the training set, and put the pose into the 3D scene as the final position of the motion according to the egocentric video. 
    We show that humans can easily solve the task with the intention clues extracted from gaze, while without the gaze information even human intelligence can be confused.}
    \label{fig:sup human}
 \end{figure}
\begin{figure}[t!]
    \centering
    \includegraphics[width=\linewidth]{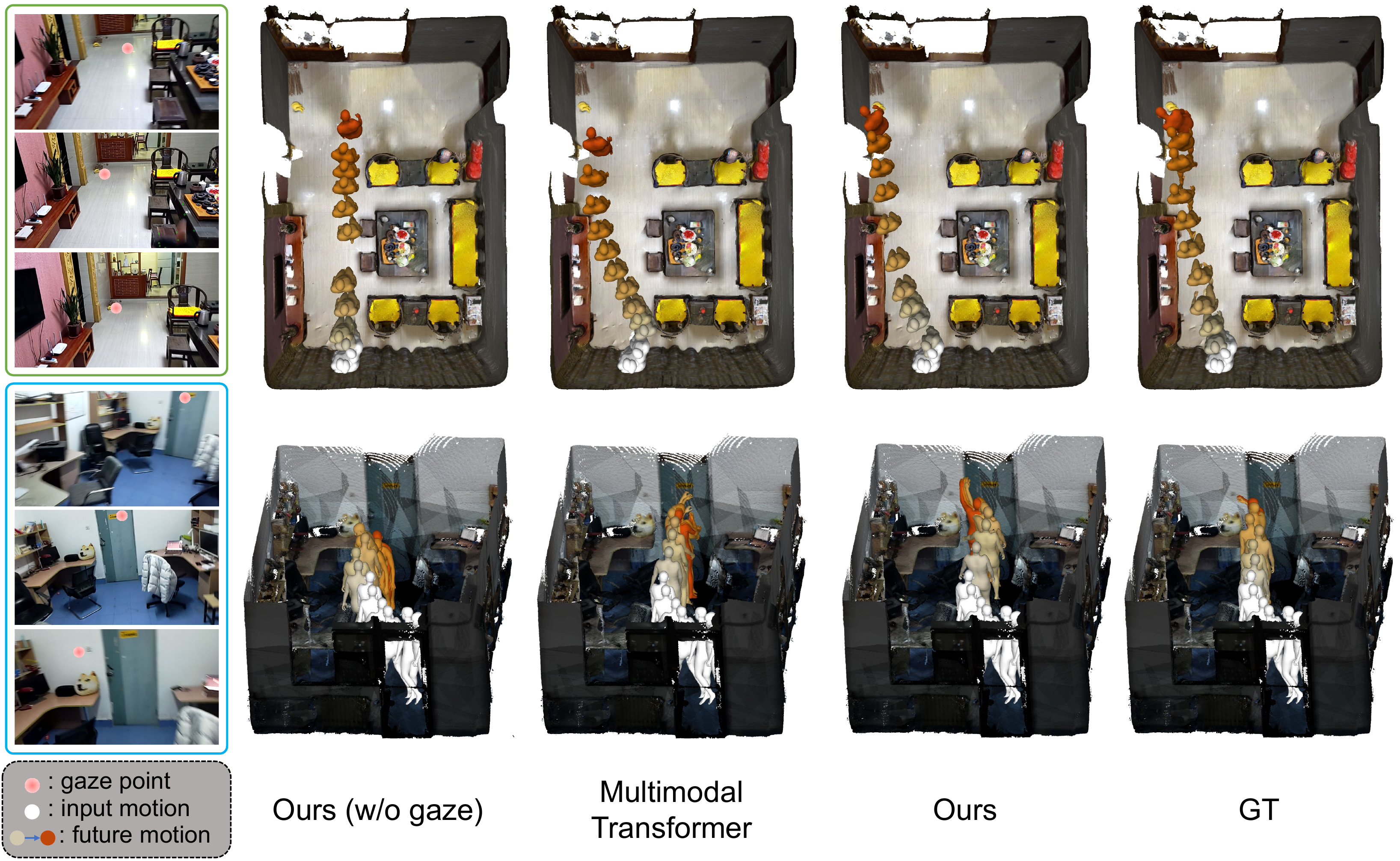}
    \caption{More Qualitative results.}
    \label{fig:sup results}
 \end{figure}
\begin{figure}[t!]
    \centering
    \includegraphics[width=\linewidth]{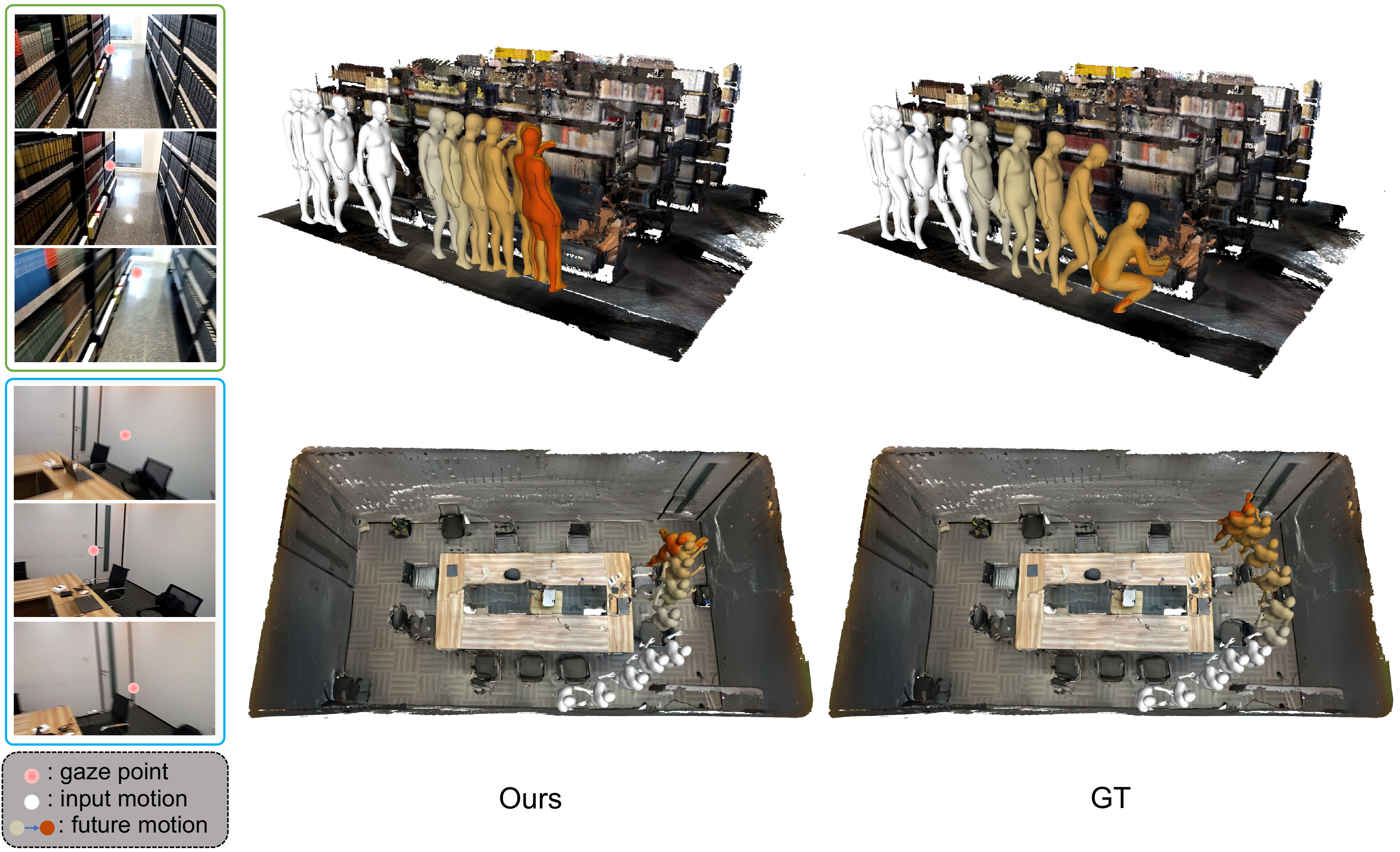}
    \caption{Failure cases of our method. When the noisy gazes account for a large portion of the input, our method is confused to interpret the subject's intention.}
    \label{fig:sup failure}
 \end{figure}
\section{GIMO Dataset}
Our dataset consists of 217 motion trajectories collected in 19 scenes by 11 subjects. 
Fig.~\ref{fig:scenes} provides an overview of the scanned scenes in our dataset, which cover a wide range of daily indoor environments, including living rooms, meeting rooms, library, lab, etc. 
Fig.~\ref{fig:sub} shows the recruited subjects collecting data in the scenes. More motion trajectories are demonstrated in Fig.~\ref{fig:sup motion}. For better visualization, please refer to the supplementary video.

\section{More Results}
\subsection{Human Evaluation}
We conduct a human evaluation experiment to validate the function of gaze in disambiguating future motion prediction. 
For simplification, the subjects predict the final pose of the motion instead of the full motion trajectory.
To this end, two human subjects are recruited and required to watch an ego-centric video (without gaze or with gaze) and infer the final pose of the trajectory.
The subjects first choose a pose from a pose database which is constructed by poses from the training set, and then put the pose into the 3D scene as the final position of the motion according to the ego-centric video they have seen. 
Fig.~\ref{fig:sup human} shows that humans can easily extract the intention clues from the gaze and solve the problem accurately, while without the gaze information even human intelligence can be confused.
\subsection{More Results of Baselines and Failure cases}
Fig.~\ref{fig:sup results} provides more results of the baseline methods, further demonstrating the superiority of our method in predicting future motion from the multi-modal gaze, motion and scene information. 
However, we find that when the input gazes are quite noisy which convey little intention clues, our method can fail to interpret the subject's goal and generate inaccurate results, as shown in~\ref{fig:sup failure}. 
Since our method predicts future motion from sparse inputs (2fps), the uninformative gazes can account for a large portion of the input. 
The problem might be mitigated by leveraging high fps inputs since we find that in the recorded sequences the most attention is paid to objects related to the destination of the motion.

\end{document}